\documentclass{article}

\usepackage{arxiv}

\usepackage[utf8]{inputenc}
\usepackage[T1]{fontenc}
\usepackage[round,authoryear]{natbib}
\usepackage{url}
\usepackage{booktabs}
\usepackage{nicefrac}
\usepackage{microtype}
\usepackage{graphicx}
\usepackage[normalem]{ulem}
\usepackage{xcolor}
\usepackage{algorithm}
\usepackage{algorithmicx}
\usepackage{algpseudocode}
\usepackage{float}
\usepackage{multirow}
\usepackage{comment}
\usepackage{caption}
\usepackage{wrapfig}
\usepackage{pifont}
\usepackage{etoolbox}
\usepackage{placeins}
\usepackage{hyperref}

\usepackage{amsmath,amsfonts,bm}

\def\eqref#1{equation~\ref{#1}}

\def\1{\bm{1}}

\DeclareMathAlphabet{\mathsfit}{\encodingdefault}{\sfdefault}{m}{sl}
\SetMathAlphabet{\mathsfit}{bold}{\encodingdefault}{\sfdefault}{bx}{n}

\newcommand{\ourmethod}{\textit{OmniCache}}

\AtBeginEnvironment{algorithmic}{\footnotesize}
\algrenewcommand\alglinenumber[1]{\scriptsize #1}

\title{\ourmethod{}: Multidimensional Hierarchical Feature Caching for Diffusion Models}

\author{
\textbf{Zhaoyuan He \quad Muhammad Muaz \quad Lili Qiu}\\[0.4em]
Department of Computer Science, The University of Texas at Austin, Austin, TX, USA
}
\date{}

\begin{document}

\maketitle

\begin{abstract}
High-resolution image and video diffusion models, including SD3, FLUX, and recent video diffusion transformers, have substantially improved generative quality but remain expensive at inference time because they repeatedly evaluate attention-heavy denoisers over many sampling steps. We address this inefficiency by exploiting redundancy in intermediate diffusion features rather than changing model weights or retraining. We identify four complementary redundancy sources in image and video generation: intra-frame, inter-frame, motion, and denoising-step redundancy. Based on this analysis, we propose \ourmethod{}, a unified hierarchical caching framework that performs multidimensional feature reuse through Token Cache, Frame Cache, Block Cache, and Layered Cache. Unlike token-merging baselines that average matched features, \ourmethod{} uses similarity matching to select cacheable features, skips redundant computation, and restores positionally consistent cached activations, preserving feature order and spatial-temporal structure. The resulting framework reuses spatial features in temporal layers and temporal features in spatial layers, while Layered Cache captures cross-step redundancy at the model-layer level. Across SD3, SVD-XT, and Latte, \ourmethod{} reduces inference latency by up to 35\%, 25\%, and 28\%, respectively, while maintaining visual fidelity and motion coherence in a training-free setting.
\end{abstract}

\suppressfloats[t]
\section{Introduction}

Recent image and video diffusion systems, including Sora and Veo~\citep{SORA,DeepMindVeo}, have made high-resolution, temporally coherent generation increasingly practical. This progress comes with a substantial inference cost. Video models must repeatedly denoise many high-resolution frames, and modern diffusion transformers further amplify this cost through attention over large spatial and temporal token sets. As resolution, frame count, and sampling steps increase, reducing redundant computation becomes essential for practical deployment.

Unlike image generation, video generation must preserve both spatial detail and frame-to-frame coherence. U-Net--based video diffusion models such as Stable Video Diffusion (SVD)~\citep{SVD}, VideoCrafter~\citep{chen2024videocrafter2}, MicroCinema~\citep{wang2023microcinema}, and AnimateDiff~\citep{guo2023animatediff} extend image denoisers with temporal layers. Diffusion transformer models such as Latte~\citep{ma2024latte} and Open-Sora~\citep{Open-Sora} alternate spatial and temporal attention. In both families, generating $N$ frames requires repeated processing across time, while increasing spatial resolution sharply increases the token count. However, natural videos are highly redundant: adjacent frames often share appearance, nearby spatial regions often share motion, and diffusion features evolve smoothly across neighboring denoising steps. The bitrate example in Table~\ref{tab:bitrate} provides a simple codec-side intuition for this redundancy.

\begin{table}[tbp]
    \centering
    \resizebox{0.5\textwidth}{!}{
         \begin{tabular}{|c|c|}
            \hline
            \textbf{Resolution \& Frame Rate} & \textbf{Bit Rate (Kbps)} \\
            \hline
            640\ensuremath{\times}360, 1 FPS & 192 \\
            \hline
            640\ensuremath{\times}360, 30 FPS & 576 \\
            \hline
            1920\ensuremath{\times}1080, 30 FPS & 2048 \\
            \hline
        \end{tabular}
    }
    \caption{Approximate H.265 recommended bit rates for representative resolutions and frame rates~\citep{hikvision2024}. The sublinear increase in bitrate relative to frame rate and pixel count illustrates the redundancy present in natural video.}
    \label{tab:bitrate}
\end{table}

We categorize the inefficiencies in diffusion-based video generation into four complementary redundancy types:

\begin{itemize}
    \item{\textit{Intra-frame redundancy}:} Repeated patterns within a frame create similar spatial tokens. Selecting representative tokens and reusing cached activations can reduce spatial attention cost.
    \item{\textit{Inter-frame redundancy}:} Adjacent frames often contain similar objects and backgrounds. Reusing spatial-layer features across similar frames can reduce repeated appearance computation while preserving frame count.
    \item{\textit{Motion redundancy}:} Motion patterns are often locally coherent across neighboring spatial regions. Reusing temporal-layer features across similar spatio-temporal blocks can reduce redundant motion computation.
    \item{\textit{Denoising-step redundancy}:} Intermediate diffusion features change gradually across neighboring sampling steps. Caching selected layer outputs across steps can reduce repeated denoiser evaluations.
\end{itemize}

Prior work has exploited parts of this structure. DeepCache~\citep{ma2023deepcacheacceleratingdiffusionmodels} and block caching~\citep{wimbauer2024cache} reuse U-Net features across denoising steps, while $\Delta$-DiT~\citep{chen2024deltadittrainingfreeaccelerationmethod}, ToCa~\citep{zou2025acceleratingdiffusiontransformerstokenwise}, Learning-to-Cache~\citep{ma2024learning}, and recent video-DiT accelerators~\citep{zhao2024realtimepab,lv2024fastercache,kahatapitiya2024adacache,liu2024teacache,zhang2025blockdancereusestructurallysimilar,liu2025taylorseer,yu2025abcache,zhou2025easycache,son2026relationalfeature,zou2026disca} extend caching, broadcasting, forecasting, or token reuse to transformer-based diffusion models. Separately, ToMe and ToMeSD~\citep{tome,tomesd} reduce transformer computation by matching and merging similar tokens through feature averaging. These methods demonstrate the value of redundancy-aware inference, but they typically focus on one reuse axis at a time and do not jointly coordinate token, frame, block, layer, and timestep decisions.

\begin{figure}[tbp]
    \centering
    \includegraphics[width=0.785\linewidth]{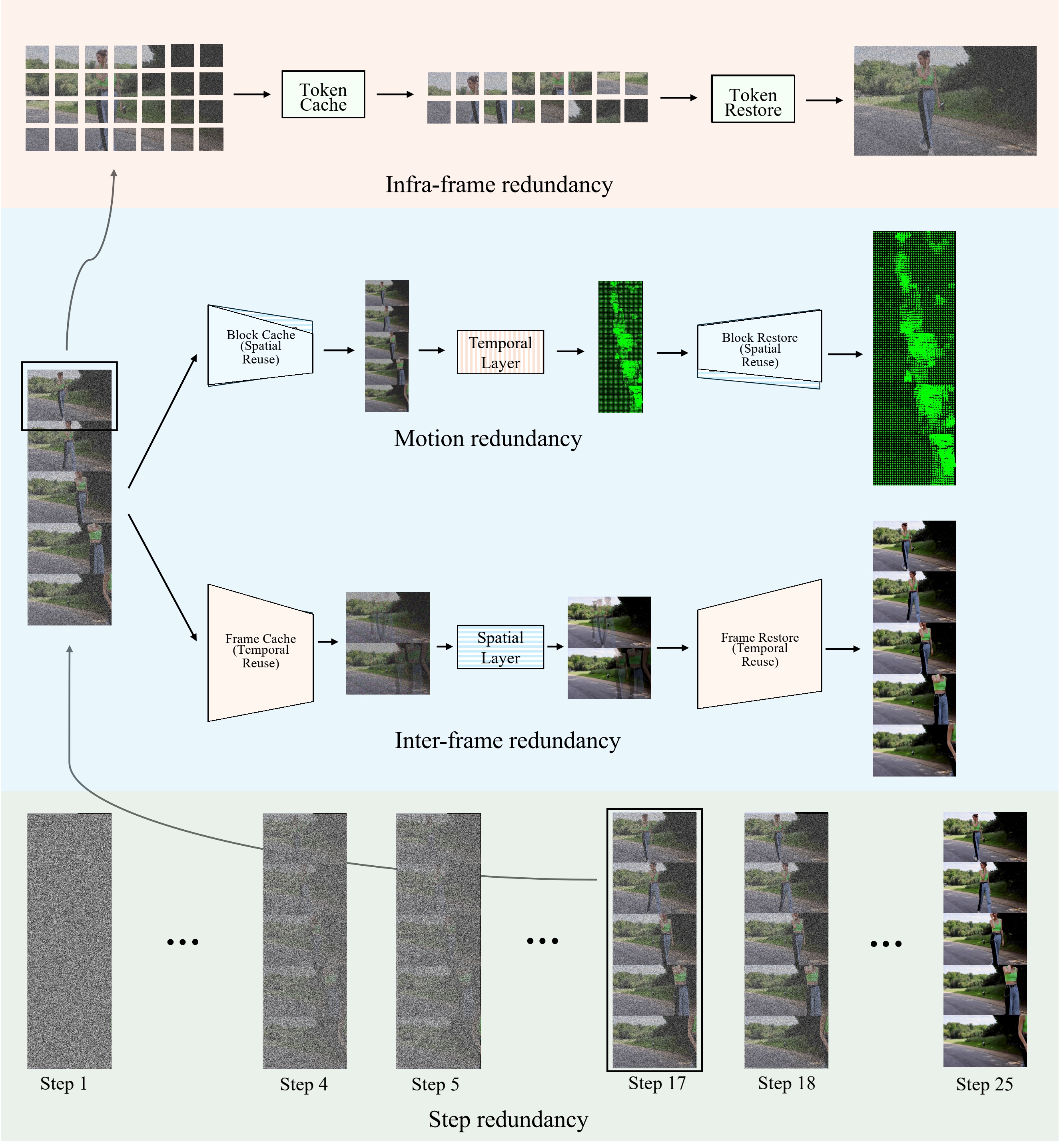}
    \caption{\ourmethod{} reduces redundancy in a hierarchical cascade across time and space. It caches features across consecutive denoising steps, exploits motion and inter-frame redundancy for non-cached features, and then identifies intra-frame redundancy among spatial tokens.}
    \label{fig:teaser}
    \vspace{-5pt}
\end{figure}

We propose \ourmethod{}, a unified hierarchical caching framework for training-free acceleration of image and video diffusion models. As shown in Figure~\ref{fig:teaser}, \ourmethod{} integrates Token Cache, Frame Cache, and Block Cache to target intra-frame, inter-frame, and motion redundancy, respectively, and combines them with a complementary Layered Cache for denoising-step redundancy. Importantly, \ourmethod{} is derived from the token-matching intuition behind token merging, but it does not average matched features. Throughout the paper, we reserve ``token merging'' for ToMe-style baselines; \ourmethod{} instead uses similarity matching to select cacheable features, skips redundant computation, and restores skipped positions from cached activations. This cache-based design preserves token order, spatial-temporal layout, and compatibility with position-sensitive video diffusion models.

We formulate caching as an adaptive resource-allocation problem: given redundancy patterns across spatial, temporal, token, layer, and denoising-step dimensions, determine where feature reuse yields the largest efficiency gain under a fixed quality budget. By combining lightweight similarity estimation, patch-level hierarchical caching, layer/timestep-aware scheduling, and GPU-optimized Triton kernels, \ourmethod{} translates redundancy into consistent wall-clock acceleration.

The core contributions of \ourmethod{} are:

\begin{enumerate}
    \item \textit{A redundancy-first formulation for diffusion inference.} We characterize inefficiencies in image and video diffusion along intra-frame, inter-frame, motion, and denoising-step axes, and formulate cache-based feature reuse as a budgeted allocation problem across these dimensions.
    \item \textit{A unified multi-granularity hierarchical caching framework.} We integrate Frame Cache, Block Cache, Token Cache, and Layered Cache, with inference-time controls over caching ratios, layer scope, and cache interval to support flexible quality--efficiency tradeoffs.
    \item \textit{Structure-aware reuse aligned with spatial-temporal layer roles.} We cache spatial information within temporal layers and temporal information within spatial layers, preserving motion and appearance better than mismatched S2S/T2T reuse.
    \item \textit{A system design for practical GPU acceleration.} We use patch-wise hierarchy, data-centric cache scheduling, and Triton gather--scatter kernels to reduce both model computation and caching overhead, achieving 25\%--35\% inference speedups on SVD-XT, Latte, and SD3 with minimal quality degradation.
\end{enumerate}

\section{Related Work}

\subsection{Diffusion Models}
Diffusion models~\citep{ho2020denoising,dhariwal2021diffusion} have become a dominant approach for high-quality image and video synthesis. Early latent diffusion systems primarily used U-Net denoisers, while Diffusion Transformers (DiT)~\citep{peebles2023scalable} replace convolutional backbones with transformer blocks and scale effectively to high-resolution generation. Recent text-to-image models such as Stable Diffusion 3~\citep{esser2024scalingrectifiedflowtransformers} and FLUX~\citep{flux2025} illustrate this shift, but also inherit the quadratic cost of attention.

Video diffusion models introduce an additional temporal dimension. Stable Video Diffusion~\citep{SVD} extends latent image diffusion with temporal modules for image-to-video generation, while SV3D~\citep{voleti2024sv3dnovelmultiviewsynthesis} adapts latent video diffusion for multi-view synthesis. Latte~\citep{ma2024latte}, Sora~\citep{SORA}, and Open-Sora~\citep{Open-Sora} further demonstrate the promise of transformer-based video diffusion. These advances motivate inference-time acceleration methods that can reduce cost without retraining or degrading temporal coherence.

\subsection{Efficient Diffusion Inference}
Efficient diffusion inference has been studied from several angles. Solver and distillation methods reduce the number of sampling steps~\citep{salimans2022progressive,lu2022dpmsolverfastodesolver,dpmplusplus,karras2022elucidatingdesignspacediffusionbased}, while caching methods reduce the cost of each denoising step. DeepCache~\citep{ma2023deepcacheacceleratingdiffusionmodels} observes that high-level U-Net features change slowly across timesteps and can be reused. Block caching~\citep{wimbauer2024cache} similarly reuses intermediate blocks to accelerate diffusion models. For diffusion transformers, $\Delta$-DiT~\citep{chen2024deltadittrainingfreeaccelerationmethod} caches feature offsets, ToCa~\citep{zou2025acceleratingdiffusiontransformerstokenwise} performs token-wise feature caching, and Learning-to-Cache~\citep{ma2024learning} learns timestep-dependent layer caching policies.

Recent work has expanded this direction for video and DiT inference. Pyramid Attention Broadcast (PAB)~\citep{zhao2024realtimepab} broadcasts attention outputs across selected timesteps for real-time video generation. FasterCache~\citep{lv2024fastercache} accelerates video diffusion with training-free feature reuse. AdaCache~\citep{kahatapitiya2024adacache} adaptively schedules caching for diffusion transformers, and TeaCache~\citep{liu2024teacache} uses timestep embeddings as a signal for when video diffusion features can be reused. The 2025--2026 literature increasingly moves beyond direct reuse toward forecasting, adaptive control, and correction: TaylorSeer~\citep{liu2025taylorseer} predicts future features with Taylor expansion, AB-Cache~\citep{yu2025abcache} connects cached feature reuse to Adams--Bashforth updates, EasyCache~\citep{zhou2025easycache} adapts caching online for large video models, and Relational Feature Caching~\citep{son2026relationalfeature} uses input--output relationships to reduce forecasting error. Very recent correction-, control-, and error-bounded variants, including AdaCorrection~\citep{liu2026adacorrection}, SoftCap~\citep{zhang2026softcap}, SpectralCache~\citep{li2026spectralcache}, and DisCa~\citep{zou2026disca}, further highlight the importance of controlling approximation error during cached diffusion inference.

\subsection{Token Matching, Merging, and Video Redundancy}
Token merging is another line of efficient transformer inference. ToMe~\citep{tome} introduces bipartite token matching to merge similar tokens in vision transformers, and ToMeSD~\citep{tomesd} adapts this idea to Stable Diffusion by inserting merge and unmerge operations around attention layers. These methods are important baselines for our work: they show that lightweight similarity matching can identify redundancy, but their feature-averaging merge can disturb token order and positional consistency in modern video diffusion models.

Several video-specific methods exploit appearance-motion structure. CMD~\citep{Yu2024EfficientVD} decomposes video generation into a content frame and low-dimensional motion latents, while LFDM~\citep{ni2023conditional} predicts latent optical flow to animate an input image. VidToMe~\citep{li2024vidtome} applies token merging to video editing, and BlockDance~\citep{zhang2025blockdancereusestructurallysimilar} reuses structurally similar spatio-temporal blocks in diffusion transformers. These works confirm that video redundancy is substantial, but each targets a relatively specific reuse axis or architecture.

In contrast, \ourmethod{} coordinates feature reuse across frame, block, token, layer, and denoising-step levels within one training-free hierarchy. It inherits the matching intuition of token merging, but replaces averaging-based merge/unmerge with cache-based selection and restoration, allowing the method to preserve spatial-temporal order while flexibly allocating a caching budget across image and video diffusion models. These properties make \ourmethod{} complementary to recent forecasting and correction methods, which primarily improve how cached features are approximated once a reuse opportunity has been selected.

\section{Methods}

\subsection{Background}

\paragraph{Diffusion Models.}
Diffusion models generate data by iteratively denoising Gaussian noise through a learned denoiser. Given a timestep $t\in\{1,\ldots,T\}$ and original image or video latent $x_0$, the forward process produces a noisy sample
$x_t=\sqrt{\bar{\alpha}_t}x_0+\sqrt{1-\bar{\alpha}_t}\epsilon$,
where $\epsilon\sim\mathcal{N}(0,I)$ and $\bar{\alpha}_t$ follows a predefined noise schedule. A noise prediction network $\epsilon_\theta(x_t,c,t)$ takes the noisy latent $x_t$, conditioning signal $c$ (e.g., text or a reference image), and timestep $t$, and is trained to predict $\epsilon$ by minimizing the standard denoising objective~\citep{ho2020denoising},
$\mathbb{E}\big[\|\epsilon-\epsilon_\theta(x_t,c,t)\|_2^2\big]$.

During inference, sampling starts from Gaussian noise $x_T$ and repeatedly applies a numerical solver to obtain $x_0$. For DDPM-style sampling~\citep{ho2020denoising}, the update can be written as
$x_{t-1}=\frac{1}{\sqrt{\alpha_t}}\!\left(x_t-\frac{\beta_t}{\sqrt{1-\bar{\alpha}_t}}\epsilon_\theta(x_t,c,t)\right)+\sigma_t z$,
with $z\sim\mathcal{N}(0,I)$. Modern solvers~\citep{song2020denoising,lu2022dpmsolverfastodesolver,dpmplusplus,karras2022elucidatingdesignspacediffusionbased} vary in formulation, but all require repeated denoiser evaluations across timesteps. Video diffusion models additionally need frame-to-frame coherence, so they commonly alternate spatial and temporal attention within U-Net or transformer blocks.

\paragraph{Token Merging Baselines and Cache-Based Restoration.}
Token Merging (ToMe)~\citep{tome} improves vision transformer efficiency by merging redundant tokens identified through bipartite similarity matching. ToMeSD~\citep{tomesd} adapts this idea to diffusion models by inserting merge and unmerge operations around attention and feed-forward layers, reducing intermediate computation while preserving the output feature-map resolution.

ToMeSD partitions tokens into source and destination sets, matches similar pairs by cosine similarity, and averages selected token pairs. During unmerging, the averaged feature is broadcast back to the original token positions. This is effective for image diffusion, but averaging and restoration can disturb token order and position-specific information, which is problematic for video diffusion models with order-sensitive positional encodings and tight spatial-temporal consistency requirements.

\ourmethod{} keeps the lightweight matching idea but replaces averaging-based merging with cache-based selection and restoration. Token Cache identifies redundant tokens, deterministically retains representative tokens for the target layer, skips computation for matched redundant positions, and restores those positions from cached activations from previous denoising steps. This distinction is central: ToMe-style methods merge features by averaging, whereas \ourmethod{} reuses positionally consistent cached features.

\begin{figure}[tbp]
    \centering
    \includegraphics[width=\linewidth]{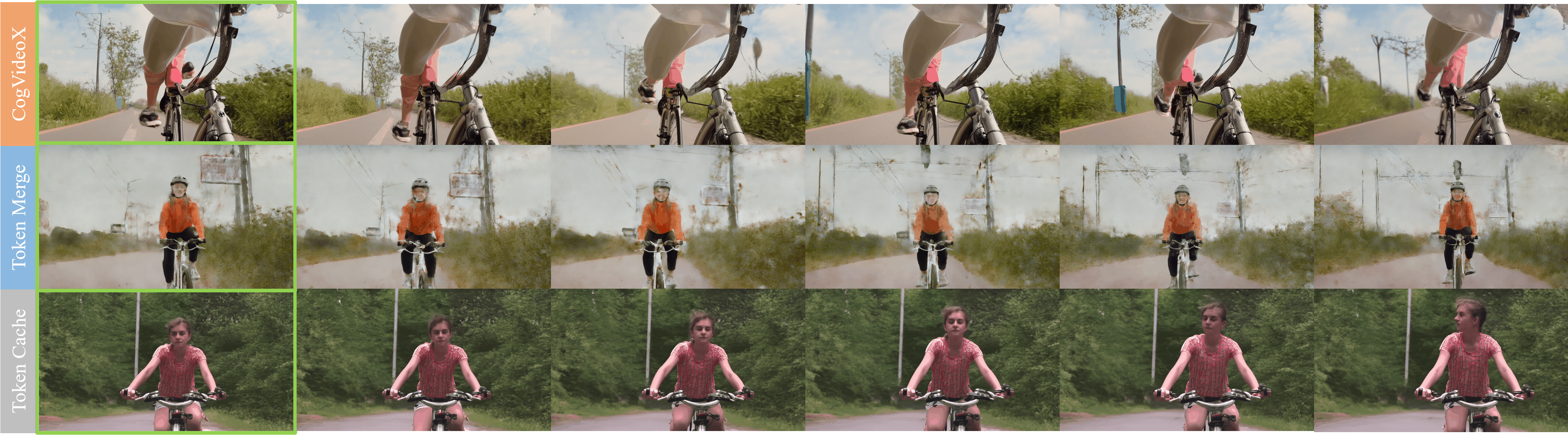}
    \caption{Limitation of averaging-based token merging in video diffusion. Cache-based restoration preserves positional consistency more effectively than broadcasting averaged features back to matched token positions.}
    \label{fig:cogvideox}
\end{figure}

\subsection{\ourmethod}

We present \ourmethod{}, a unified hierarchical caching framework that removes redundancy across multiple levels of diffusion models. As shown in Figure~\ref{fig:teaser}, it integrates Token Cache, Frame Cache, and Block Cache to address intra-frame, inter-frame, and motion redundancy, respectively, together with Layered Cache for cross-step reuse. All cache modules follow the same principle: reuse intermediate features where redundancy is high and approximation tolerance is large under an explicit inference-time budget. Rather than making independent caching decisions, \ourmethod{} treats feature reuse as a hierarchical resource-allocation problem across space, time, tokens, layers, and denoising steps.

\begin{figure}[tbp]
    \centering
    \includegraphics[width=0.6\linewidth]{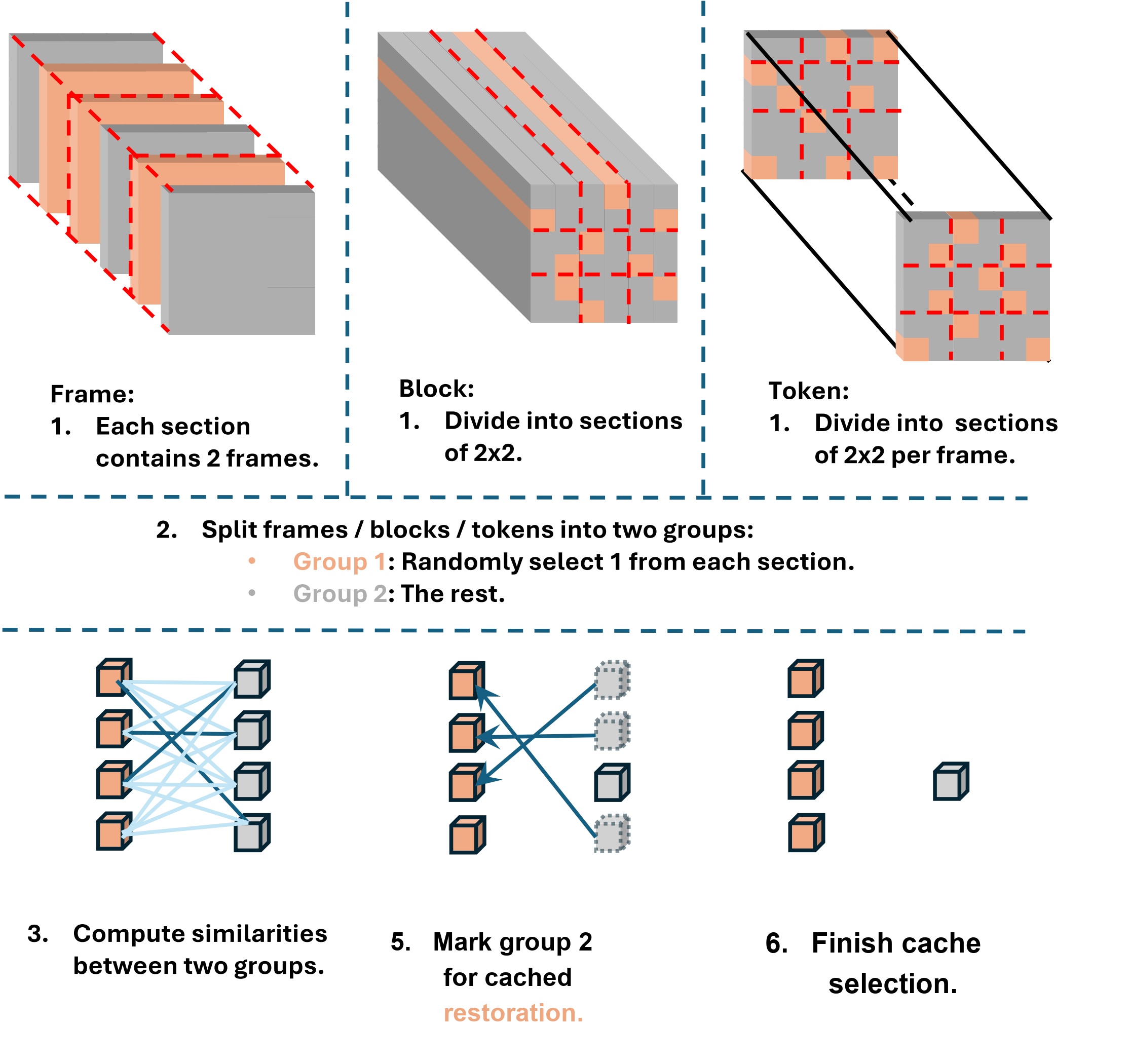}
    \vspace{-5pt}
    \caption{Illustration of \ourmethod{}'s \textit{Frame Cache}, \textit{Block Cache}, and \textit{Token Cache}. Each module uses similarity matching to select cacheable features, skips redundant computation, and restores cached activations at their original positions.}
    \label{fig:3merge}
\end{figure}

\subsubsection{Motion Redundancy and Block Cache}
Video diffusion models typically extract motion features with temporal transformer layers. For an input feature map of shape \texttt{(B, T, C, H, W)}, the temporal layer can be viewed as processing \texttt{H*W} sequences of length \texttt{T}, one for each spatial location. We call each \texttt{1*1*T} temporal-attention input a block. Many videos contain repeated or locally coherent motion, such as camera panning or rigid object movement, suggesting that temporal computation across neighboring spatial blocks can be reused.

To identify redundant motion patterns, we reshape the input \texttt{(B, T, C, H, W)} into \texttt{(B, H*W, T*C)}, isolating \texttt{H*W} block descriptors as shown in Figure~\ref{fig:3merge}. We partition the blocks into local \texttt{2*2} groups. Within each group, candidate blocks are compared by cosine similarity, and the $r_b$ most redundant blocks are marked for cached restoration while the retained representatives are computed normally.

For each denoising step $t$, Block Cache skips temporal-attention computation for the selected redundant blocks and reuses their cached activations from a previous step. Unlike ToMeSD, which averages matched features, Block Cache performs deterministic feature reuse and restores cached blocks to their original spatial positions after the temporal layer finishes computing non-cached blocks. The final feature map therefore preserves its original resolution, ordering, and spatial indexing while reducing redundant motion computation.

\subsubsection{Inter-frame Redundancy and Frame Cache}
Adjacent frames often share the same objects, background, and scene layout, differing only through modest motion. Prior image-to-video work such as LFDM~\citep{ni2023conditional} exploits this structure through optical-flow-like warping. Frame Cache applies the same redundancy principle inside the diffusion denoiser: if two frames have highly similar spatial content, the spatial-layer computation for one frame can be skipped and restored from cache without reducing the model's temporal input length.

For an input feature map of shape \texttt{(B, T, H, W, C)}, we reshape each frame to a descriptor of shape \texttt{(B, T, H*W*C)}. We split the $T$ frames into alternating candidate sets so that matching remains local while covering the full sequence. Bipartite matching then identifies the $r_f$ most redundant frames. These selected frames skip the current spatial-layer computation, and their positionally corresponding cached features from the previous denoising step are reused.

Frame Cache does not lower the generated frame rate. Temporal layers still receive a full-length sequence and operate over the original frame positions. Thus, spatial computation is reduced for redundant frames, while temporal processing can continue to model frame-to-frame variation and preserve motion richness.

\subsubsection{Intra-frame Redundancy and Token Cache}
After exploiting inter-frame redundancy, substantial redundancy often remains within each frame: sky, water, background texture, and other repeated regions produce similar spatial tokens. Token Cache reduces this intra-frame redundancy before spatial layers while preserving each skipped token's original position for restoration.

Token Cache consumes the same feature map and frame-pair metadata used by Frame Cache. Applying token selection only after dropping frame computations would reduce temporal diversity, so we keep full frame bookkeeping and apply token selection to both matched frame pairs and unmatched frames. For frame pairs selected by Frame Cache, we reshape the paired features from \texttt{(B, $2r_f$, H, W, C)} to \texttt{(B*$r_f$, 2*H*W, C)}. Within each local \texttt{2*2*2} cube, Token Cache retains representative tokens for the spatial layer and marks the $r_t$ most redundant positions for cached restoration. For unmatched frames, the same procedure is applied within local \texttt{2*2} spatial neighborhoods.

After the spatial layer computes the retained tokens, skipped token positions are restored from cached activations associated with the same positions at the previous denoising step. The retained and restored tokens are then scattered back to the original frame layout. This reduces spatial-token computation while preserving the original frame count and token ordering.

\subsection{Hierarchical Caching}

\begin{figure}[tbp]
    \centering
    \includegraphics[width=0.8\linewidth]{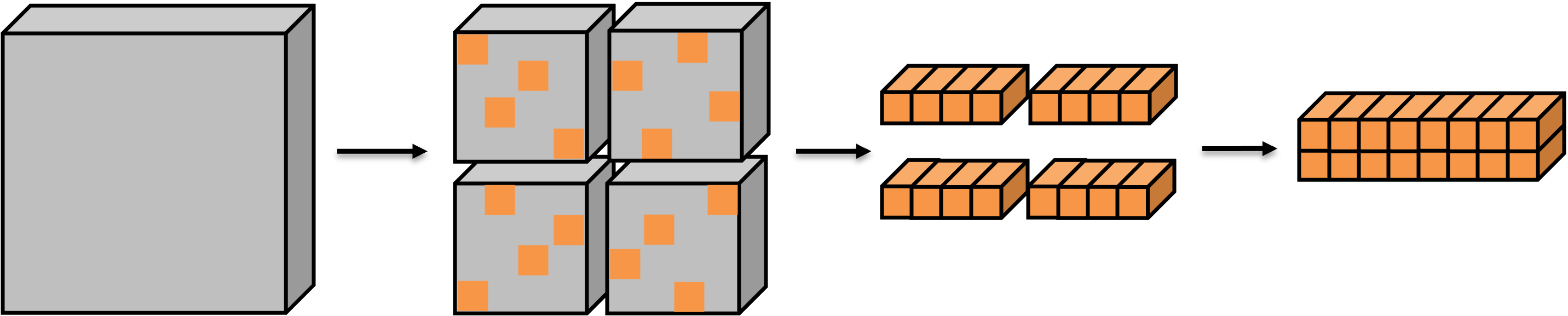}
    \caption{Illustration of Hierarchical Caching. Similarity search and feature reuse are applied locally within patches before optionally expanding to coarser spatial scopes.}
    \label{fig:hier_merge}
\end{figure}

Similarity search over full-resolution feature maps can be expensive. For Frame Cache, directly comparing descriptors of shape \texttt{(B, T, H*W*C)} is costly because spatial content is high-dimensional. We therefore divide spatial content into \texttt{k*k} patches, reshape the feature map to \texttt{(B, T, k, H/k, k, W/k, C)}, and then permute it to \texttt{(B*k*k, T, (H/k)*(W/k)*C)} for local similarity computation. Patch-local matching reduces Frame Cache overhead by more than $20\times$ on SVD-XT without degrading visual quality or motion.

The same hierarchy applies to Block Cache and Token Cache because nearby pixels and neighboring blocks are more likely to be redundant than distant ones. In SVD-XT, patch-local matching reduces Block Cache and Token Cache overhead by approximately $18\times$ and $15\times$, respectively. After patch-level caching, the patch size can be increased for coarser reuse, or patches can be restored to full frames for global matching. This hierarchy lets \ourmethod{} trade off similarity-search overhead, cache granularity, and generation quality.

\paragraph{Implementation Efficiency.}
To ensure that hierarchical caching translates into wall-clock speedups, we implement the key selection and restoration operations using custom Triton kernels. These kernels optimize gather--scatter memory access and reduce runtime overhead; detailed kernel design and profiling results are provided in Appendix~\ref{appendix:triton}.

\subsection{Step Redundancy and Layered Cache}
\label{sec:method_cache_proposal}

Prior work such as DeepCache and $\Delta$-DiT~\citep{ma2023deepcacheacceleratingdiffusionmodels,chen2024deltadittrainingfreeaccelerationmethod} shows that feature distributions within individual layers often evolve slowly across adjacent denoising steps, making cross-step feature reuse effective. Newer cache accelerators further show that future features can be forecast from prior-step trajectories~\citep{liu2025taylorseer,yu2025abcache} or corrected using input--output relationships and adaptive error estimates~\citep{son2026relationalfeature,liu2026adacorrection}. These methods motivate accurate approximation once a cached layer is chosen; our Layered Cache instead focuses on the complementary scheduling question of which layer--timestep pairs should reuse features under a target speedup and quality budget.

We formulate step-level caching as a data-centric, budget-constrained scheduling problem that explicitly balances efficiency gains against quality degradation across layers and denoising timesteps. For a given diffusion model, we profile each transformer or U-Net layer over a range of denoising steps using prompts of varying length from MS-COCO~2017. For each layer--timestep pair $(l,t)$, we measure average execution latency $L_{l,t}$ and quality degradation $Q_{l,t}$ when caching is applied at that location, using metrics such as FID, CLIP, or FVD depending on the task. We then assign each pair a score
$w_{l,t}=Q_{l,t}/L_{l,t}$,
which represents quality loss per unit of saved computation.

Given a target speedup $s_r$ and cache interval $N$, Layered Cache greedily selects low-score layer--timestep pairs until the accumulated saved latency reaches the required budget. This provides a knapsack-style approximation that prioritizes cache placements with favorable quality--efficiency tradeoffs. The resulting cache proposal is summarized in Algorithm~\ref{alg:layer_cache}.

To avoid redundant or invalid cache operations, we refine the initial schedule with the model's computation graph. By tracking dependencies between cached outputs across layers and timesteps, the refinement step removes unnecessary cache operations while preserving execution order and data availability. Algorithm~\ref{alg:ref_alg} summarizes this process, with full details in Appendix~\ref{appendix:refinement}.

\begin{figure*}[tbp]
\centering
\begin{minipage}[t]{0.47\textwidth}
\begin{algorithm}[H]
\caption{Cache Schedule Proposal}\label{alg:layer_cache}
\begin{algorithmic}[1]
\Require candidates $\mathcal{P}$, latencies $L$, quality losses $Q$, target speedup $s_r$, cache interval $N$
\Function{Get\_Cache\_Strategy}{}
    \State $w_{l,t} \gets Q_{l,t}/L_{l,t}$ for each $(l,t)\in\mathcal{P}$
    \State sort $\mathcal{P}$ by increasing $w_{l,t}$
    \State $\textsc{TargetSave} \gets (1 - 1/s_r)\sum_{(l,t)\in\mathcal{P}} L_{l,t}$
    \State $\textsc{Saved} \gets 0$; $\mathcal{C}\gets\emptyset$
    \While{$\textsc{Saved} < \textsc{TargetSave}$}
        \State $(l,t) \gets$ next candidate in sorted $\mathcal{P}$
        \State $\mathcal{C}\gets\mathcal{C}\cup\{(l,t,N)\}$
        \State $\textsc{Saved} \gets \textsc{Saved} + L_{l,t}$
    \EndWhile
    \State \Return $\mathcal{C}$
\EndFunction
\end{algorithmic}
\end{algorithm}
\end{minipage}
\hfill
\begin{minipage}[t]{0.47\textwidth}
\begin{algorithm}[H]
\caption{Cache Strategy Refinement}\label{alg:ref_alg}
\begin{algorithmic}[1]
\Require Model graph $\mathcal{G}$, CacheStrategy $\mathcal{C}$
\State \textbf{Preprocess:} Initialize metadata for nodes
\State \textbf{Build Dependencies:} Set child/parent relations
\State \textbf{Count Dependencies:} Compute child\_num, parent\_num
\State \textbf{Prune Redundancies:} Remove unnecessary cache steps
\State \textbf{Prune Consecutive Steps:} Drop identical outputs
\State \textbf{Recalculate Counters:} Update dependency counts
\State \textbf{Align Steps:} Match with parent cache intervals
\State \Return Optimized strategy $\hat{\mathcal{C}}$
\end{algorithmic}
\end{algorithm}
\end{minipage}
\caption{\textbf{(Left)} Algorithm for generating cache schedule proposals.
\textbf{(Right)} Refinement process for optimizing cache strategies.}
\end{figure*}

\FloatBarrier
\section{Experiments}

\subsection{Experimental Settings}

\textbf{Model Configurations.}
We evaluate \ourmethod{} on representative image and video diffusion models: Stable Video Diffusion XT (SVD-XT) for image-to-video generation, Latte-1 T2V for text-to-video generation, and Stable Diffusion 3 medium (SD3-medium)~\citep{esser2024scalingrectifiedflowtransformers} for text-to-image generation. All methods are applied directly to pretrained models without retraining. SD3-medium uses a Flow Matching Euler Discrete scheduler with 28 sampling steps, while SVD-XT uses an Euler Discrete sampler with 25 steps. Unless otherwise stated, SVD-XT generates 25 frames at $576\times1024$ resolution, and Latte-1 T2V generates 16 frames at $512\times512$ resolution. Runtime is measured on an NVIDIA A100 GPU with 40 GB of memory.

\textbf{Evaluation and Metrics.}
For image-to-video evaluation, we generate 10,000 videos with SVD-XT on UCF101 and report FVD. For text-to-image evaluation, we generate images from 8,000 randomly sampled MS-COCO 2014 captions and compute FID using 20,000 real images. We additionally report CLIP score for image-text alignment.

\subsection{Design Choices}
\label{sec:design_choices}

We investigate caching ratios and layer selections to identify where hierarchical caching improves efficiency without degrading quality. Across models, early and late layers are consistently more sensitive to feature reuse, aligning with prior observations that early and late diffusion blocks play important roles in coarse structure and fine detail synthesis.

For SVD-XT, the final spatial transformer layers are particularly sensitive to Frame Cache. We therefore exclude the last decoder layer from Frame Cache and avoid applying Frame Cache or Token Cache to FFN layers, which consistently degrades visual quality. Under these constraints, we observe a stable operating regime with caching ratios up to approximately $r_f \leq 50\%$, $r_b \leq 40\%$, and $r_t \leq 50\%$; more aggressive settings introduce over-smoothing and motion artifacts.

For Latte, we exclude the first and last transformer layers and apply $r_f=30\%$, $r_b=40\%$, and $r_t=20\%$ across the remaining layers. We again restrict Frame Cache and Token Cache to attention blocks. For SD3-medium, we apply hierarchical token caching to both attention and FFN layers but exclude the top and bottom four transformer blocks, which are highly sensitive to caching. This behavior is consistent with $\Delta$-DiT~\citep{chen2024deltadittrainingfreeaccelerationmethod}, which observes distinct roles for early and late DiT blocks.

Detailed quantitative results for different ratio settings are reported in Tables~\ref{tab:sd3_results} and~\ref{tab:fvd_ablation}. Based on these ablations, we recommend conservative default settings around $r_f \leq 40\%$, $r_b \leq 30\%$, and $r_t \leq 40\%$ when deploying on new models or content distributions.

\subsection{Main Results}

\begin{figure}[H]
    \centering
    \includegraphics[width=\linewidth]{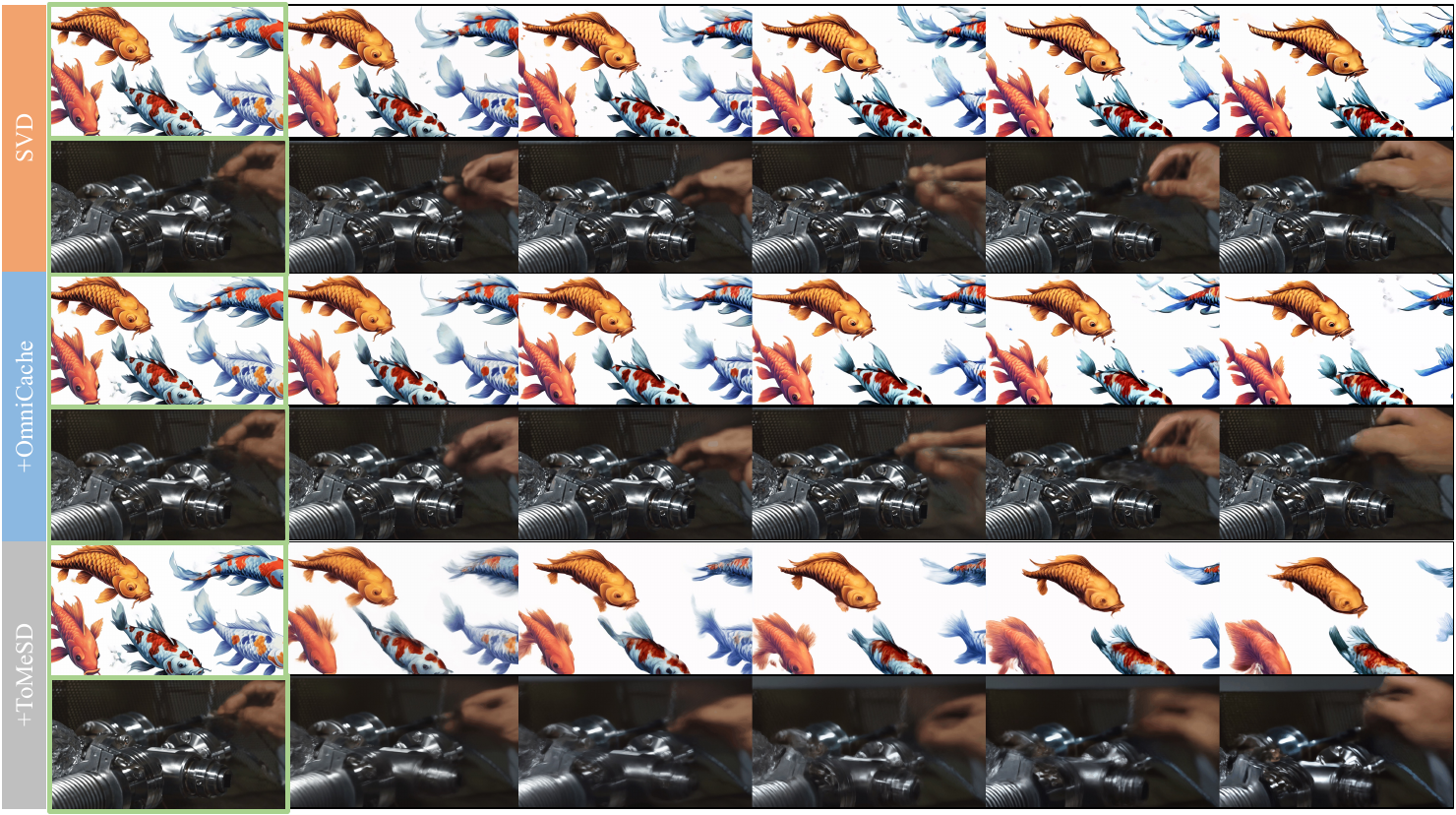}
    \caption{Performance of \ourmethod{} on SVD-XT. With $r_f$, $r_b$, and $r_t$ set to (50\%, 30\%, 50\%), \ourmethod{} is compared against ToMeSD at a comparable 65\% token-reduction setting. \ourmethod{} closely matches original SVD-XT quality while improving inference time by 25\%.}
    \label{fig:svd_main}
\end{figure}

\begin{figure}[H]
    \centering
    \includegraphics[width=\linewidth]{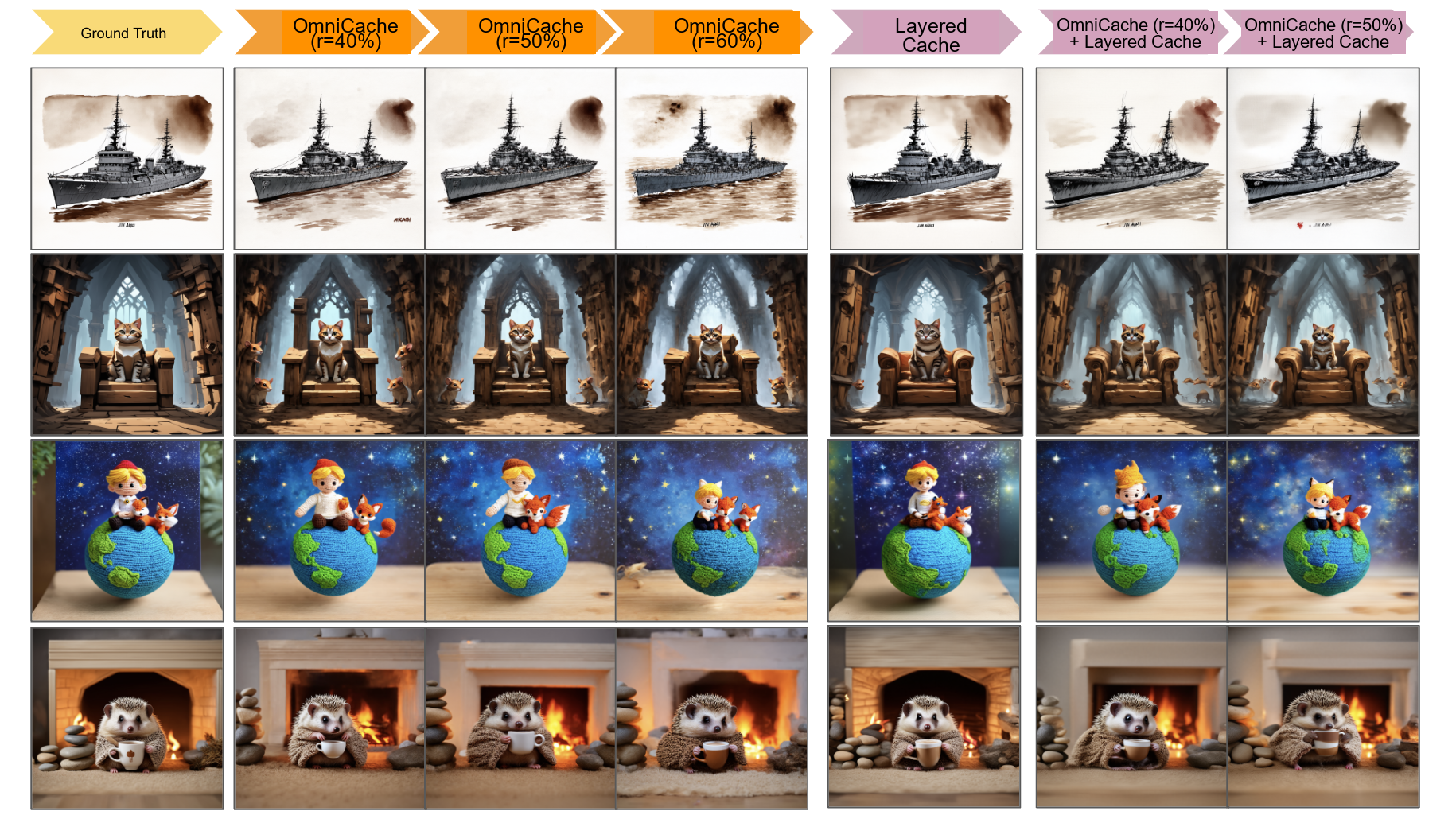}
    \caption{Visualization examples for SD3-medium with hierarchical Token Cache, Layered Cache, and their combination. Token Cache ratios of 40\%, 50\%, and 60\% are shown. Prompts used to generate these images are provided in the appendix.}
    \label{fig:sd3_main}
\end{figure}

\begin{figure}[H]
    \centering
    \includegraphics[width=0.9\linewidth]{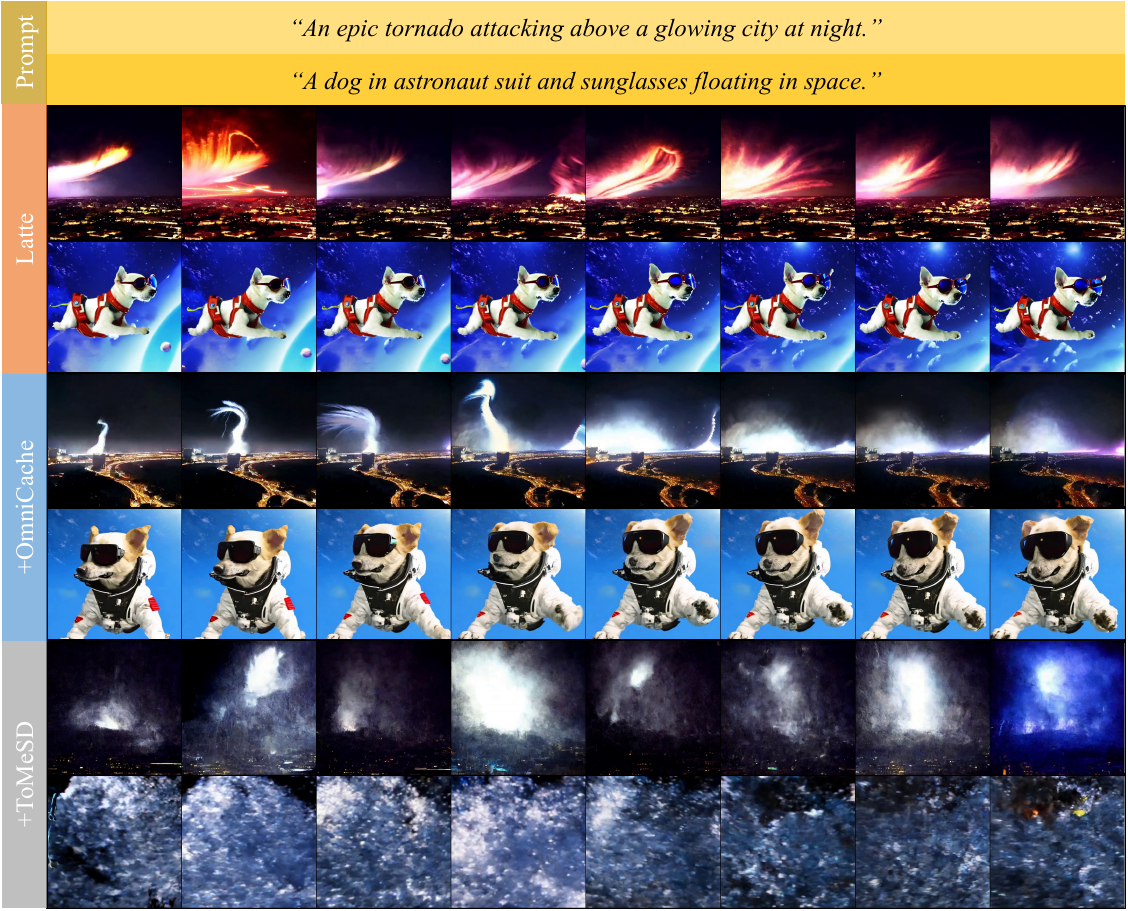}
    \caption{Performance of \ourmethod{} on Latte. With $r_f$, $r_b$, and $r_t$ set to (30\%, 40\%, 20\%), \ourmethod{} is compared against ToMeSD at a comparable 25\% token-merging setting. \ourmethod{} closely matches original Latte quality while improving inference time by 28\%.}
    \label{fig:latte_main}
\end{figure}

\paragraph{Training-free acceleration on image and video diffusion models.}
We evaluate \ourmethod{} on U-Net--based image-to-video generation (SVD-XT), DiT-based text-to-video generation (Latte), and DiT-based text-to-image generation (SD3-medium). As shown in Figures~\ref{fig:svd_main}, \ref{fig:sd3_main}, and~\ref{fig:latte_main}, \ourmethod{} preserves generation quality while reducing inference latency by 25\% on SVD-XT, 28\% on Latte, and 35\% on SD3. At comparable token-reduction settings, \ourmethod{} produces higher-quality outputs than ToMeSD-based token-merging baselines.

\newcommand{\cmark}{\ding{51}}%
\newcommand{\xmark}{\ding{55}}%

\begin{table}[H]
    \centering
    \resizebox{0.78\textwidth}{!}{
        \begin{tabular}{p{4cm}|*{3}{c}|*{2}{c}}
            \hline
            \multicolumn{1}{c|}{\textbf{Method}} & \multicolumn{3}{c|}{\textbf{Latency and Speedup}} & \multicolumn{2}{c}{\textbf{Evaluation Metrics}} \\
            \hline
            \textbf{} & \textbf{S$_\text{DiT}$} $\uparrow$ & \textbf{L$_\text{DiT}$ (ms)} & \textbf{Retrain} & \textbf{FID} $\downarrow$ & \textbf{CLIP} $\uparrow$ \\
            \hline
            SD3-medium & 1.00 & 138.36 & \xmark & 31.57 & 18.60 \\
            \hline
            \ourmethod{} 40\% & 1.20 & 114.96 & \xmark & 31.42 & 20.26 \\
            \ourmethod{} 50\% & 1.27 & 109.00 & \xmark & 31.94 & 20.32 \\
            \ourmethod{} 60\% & 1.35 & 102.88 & \xmark & 32.45 & 20.38 \\
            \hline
            Layered Cache & 1.75 & 79.00 & \xmark & 31.36 & 20.11 \\
            \hline
            \ourmethod{}+Layered Cache 40\% & 2.28 & 60.55 & \xmark & 31.14 & 20.28 \\
            \ourmethod{}+Layered Cache 50\% & 2.32 & 59.71 & \xmark & 31.95 & 20.30 \\
            \hline
        \end{tabular}
    }
    \caption{\textbf{Quantitative comparison of text-to-image generation} on MS-COCO 2014 with SD3-medium. DiT latency is reported in milliseconds.}
    \label{tab:sd3_results}
    \vspace{-5pt}
\end{table}

\paragraph{Quantitative metrics for SVD-XT.}
For SVD-XT, we adopt the best-performing quantitative configuration with $r_f=50\%$, $r_b=40\%$, and $r_t=50\%$. As shown in Table~\ref{tab:fvd_ablation}, \ourmethod{} achieves an FVD score nearly identical to the non-caching SVD-XT baseline while reducing inference latency by 25\% in a training-free manner.

For comparison, we evaluate ToMeSD using its best-performing 40\% token-merging setting in this sweep. Although ToMeSD improves efficiency, \ourmethod{} better preserves FVD and reduces runtime by an additional 11\% relative to ToMeSD, demonstrating a stronger quality--efficiency tradeoff than single-axis token merging.

\begin{table}[tbp]
    \centering
    \small
    \setlength{\tabcolsep}{6pt}
    \renewcommand{\arraystretch}{1.08}
    \begin{tabular}{cc|cc}
        \hline
        \textbf{Method} & \textbf{Ratio(s) (\%)} & \textbf{FVD $\downarrow$} & \textbf{s/video $\downarrow$} \\
        \hline
        SVD-XT (no caching) & -- & 502.68 & 110 \\
        \hline
        Token Merge (ToMeSD) & 40 & 503.8 & 92.6 \\
        \hline
        \ourmethod{} & 50/40/50 & 503.12 & 82.4 \\
        \hline
        Frame Cache & 20 & 489.14 & 105 \\
        & 40 & 495.89 & 99.3 \\
        & 60 & 535.07 & 94.8 \\
        \hline
        Token Cache & 20 & 497.49 & 102 \\
        & 40 & 502.5 & 94.3 \\
        & 60 & 517.64 & 84 \\
        \hline
        Block Cache & 10 & 499.84 & 107 \\
        & 30 & 502.71 & 103 \\
        & 50 & 528.24 & 98.8 \\
    \end{tabular}
    \caption{Quantitative evaluation for different caching ratios on SVD-XT. The \ourmethod{} ratio denotes $r_f/r_b/r_t$.}
    \label{tab:fvd_ablation}
\end{table}

\begin{figure}[H]
    \centering
    \includegraphics[width=0.96\linewidth]{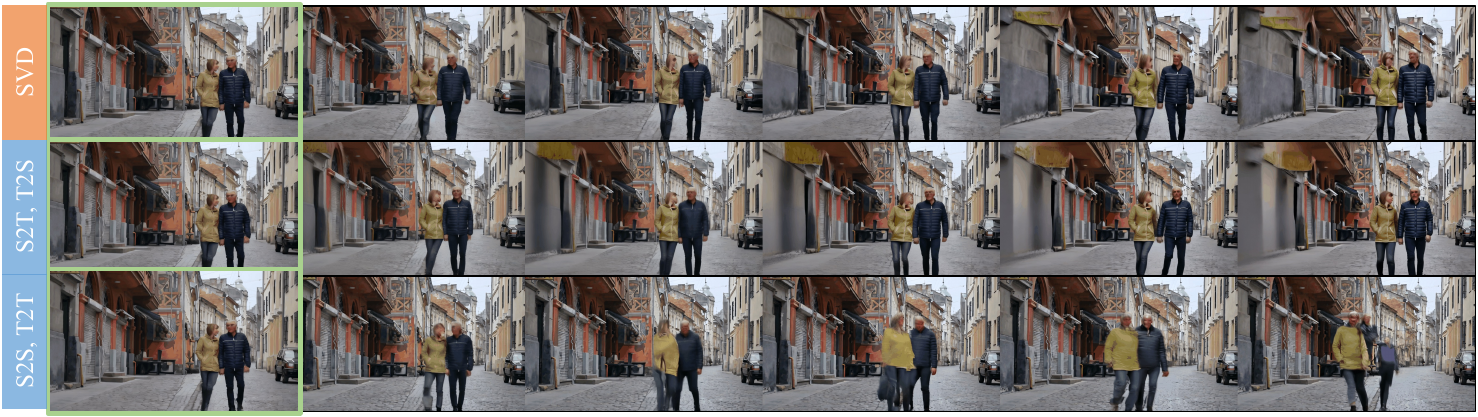}
    \caption{S2T/T2S versus S2S/T2T reuse on SVD-XT with $r_f$, $r_b$, and $r_t$ set to (30\%, 30\%, 0\%). Token Cache is omitted because this experiment isolates the placement of Frame Cache and Block Cache across spatial and temporal layers. S2T and T2S retain more information than mismatched S2S and T2T reuse.}
    \label{fig:reverse}
\end{figure}

\paragraph{Structure-aware spatial and temporal reuse.}
Figure~\ref{fig:teaser} suggests that spatial information should be reused in temporal layers, while temporal information should be reused in spatial layers. We test this hypothesis by swapping the locations of Frame Cache and Block Cache, with both ratios set to 30\%. As shown in Figure~\ref{fig:reverse}, Spatial-to-Temporal reuse (S2T) and Temporal-to-Spatial reuse (T2S) closely resemble the original SVD-XT output. In contrast, Spatial-to-Spatial reuse (S2S) and Temporal-to-Temporal reuse (T2T) cause visible blur, artifacts, or degraded motion. This supports the structure-aware design of \ourmethod{}.

\begin{figure}[H]
    \centering
    \includegraphics[width=0.96\linewidth]{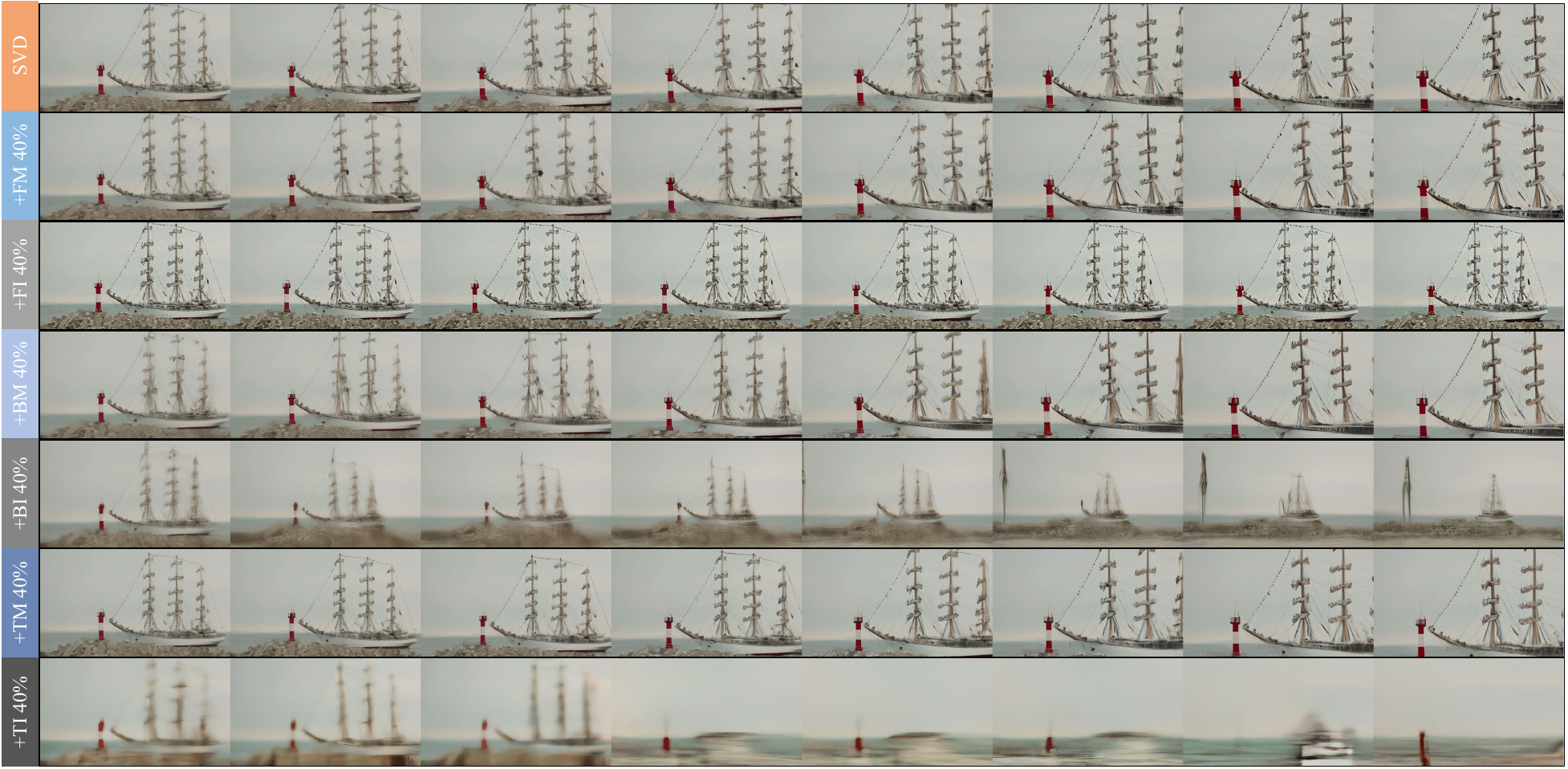}
    \caption{Comparison of Frame Cache, Block Cache, and Token Cache with frame interpolation (FI), block interpolation (BI), and token interpolation (TI). All feature-reduction ratios are set to 40\%, and experiments are conducted on SVD-XT.}
    \label{fig:mvsi}
\end{figure}

\paragraph{\ourmethod{} enables efficient feature reuse.}
We compare Frame Cache, Block Cache, and Token Cache against direct interpolation baselines at the same 40\% feature-reduction ratio. As shown in Figure~\ref{fig:mvsi}, Frame Interpolation (FI) significantly reduces perceived frame rate, Block Interpolation (BI) yields blurry results and degraded motion fidelity, and Token Interpolation (TI) causes spatial blur. In contrast, all three caching modules in \ourmethod{} produce results that closely match original SVD-XT. These results show that cache-based restoration exploits redundancy more effectively than direct feature interpolation.

\FloatBarrier

\subsection{Ablations}

\paragraph{Caching ratios.}
Table~\ref{tab:fvd_ablation} analyzes the effect of individual Frame Cache, Block Cache, and Token Cache ratios on SVD-XT. We observe little to no FVD degradation up to 40\% frame caching, 40\% token caching, and 30\% block caching. Beyond these ranges, quality begins to degrade, indicating that conservative caching ratios provide a reliable operating regime.

\paragraph{Layered Cache speedup and quality.}
Table~\ref{tab:sd3_layer_cache_ablation} summarizes Layered Cache schedules for different target speedups on SD3-medium. We set the cache interval to $N=5$ and compare against DeepCache-style intervals ($N=2,3$). End-to-end speedup is measured directly, while FID and CLIP measure perceptual quality and LPIPS, SSIM, and PSNR measure pixel-level similarity on 2,000 random MS-COCO 2017 image--text pairs. This diagnostic subset differs from the MS-COCO 2014 evaluation used in Table~\ref{tab:sd3_results}, so absolute FID and CLIP values should not be compared across the two tables.

\begin{table}[tbp]
\centering
\small
\setlength{\tabcolsep}{5pt}
\renewcommand{\arraystretch}{1.05}
\resizebox{0.82\textwidth}{!}{
\begin{tabular}{lcccccc}
\toprule
\multirow{2}{*}{\textbf{Method}} & \multirow{2}{*}{\textbf{Speedup}} &
\multicolumn{2}{c}{\textbf{Perceptual}} &
\multicolumn{3}{c}{\textbf{Pixel-wise}} \\
\cmidrule(lr){3-4} \cmidrule(lr){5-7}
 &  & \textbf{FID $\downarrow$} & \textbf{CLIP $\uparrow$} &
 \textbf{LPIPS $\downarrow$} & \textbf{SSIM $\uparrow$} & \textbf{PSNR $\uparrow$} \\
\midrule
\textbf{SD3 (w/o caching)} & 1.000x & 39.187  & 32.169 & 0.000 & 1.000 & $\infty$ \\
\midrule
\textbf{DeepCache} (int=2) & 1.8928x & 39.0282 & --     & --    & --    & --  \\
DeepCache (int=3) & 2.5552x & 39.6342 & --     & --    & --    & --  \\
\midrule
\textbf{Our Method} & 1.645x  & 37.546  & 32.096 & 0.329 & 0.634 & 14.801 \\
                    & 1.811x  & 37.222  & 32.250 & 0.279 & 0.681 & 16.333 \\
                    & 1.889x  & 36.598  & 32.027 & 0.367 & 0.597 & 13.955 \\
                    & 1.969x  & 36.852  & 31.993 & 0.367 & 0.598 & 13.968 \\
                    & 2.184x  & 35.353  & 31.676 & 0.363 & 0.596 & 14.680 \\
                    & 2.348x  & 35.560  & 31.579 & 0.362 & 0.596 & 14.617 \\
                    & 2.506x  & 35.619  & 31.676 & 0.346 & 0.615 & 15.053 \\
                    & 2.551x  & 35.979  & 31.511 & 0.364 & 0.596 & 14.672 \\
                    & 2.655x  & 35.991  & 31.207 & 0.374 & 0.584 & 14.626 \\
                    & 2.778x  & 36.202  & 31.178 & 0.376 & 0.583 & 14.597 \\
\bottomrule
\end{tabular}
}
\caption{Performance metrics of Stable-Diffusion-3 for different speedups. Image metrics are computed using 2,000 random image--text pairs from the MS-COCO 2017 dataset.}
\label{tab:sd3_layer_cache_ablation}
\vspace{-5pt}
\end{table}

\FloatBarrier
\section{Limitations}

\ourmethod{} is a training-free inference-time acceleration method, so its effectiveness depends on redundancy already present in the model's intermediate features. When redundancy is limited--for example, videos with extremely fast non-repetitive motion, dense independent object movement, or prompts requiring globally changing fine detail in every frame--aggressive caching can cause over-smoothing or degraded motion fidelity. Our ablations identify conservative operating regimes where speedup improves while perceptual quality remains stable, but the best caching ratio may still depend on the target model and content distribution.

The cache schedule used by \ourmethod{} is budget-aware but heuristic. We formulate cache placement as a benefit-cost tradeoff and refine the resulting schedule with graph dependencies, yet we do not claim global optimality. More adaptive strategies, such as prompt-conditioned policies, reinforcement learning, or black-box optimization, could further improve robustness by selecting caching ratios and intervals per input.

\ourmethod{} is orthogonal to many other diffusion acceleration techniques. Step-reduction and distillation methods reduce the number of denoising steps, while \ourmethod{} reduces the computation within each step; combining them could yield multiplicative speedups. Similarly, quantization and optimized attention kernels reduce the cost of individual operations, whereas \ourmethod{} reduces how many operations need to be executed. These combinations may require conservative cache settings because changes in numerical precision or solver trajectories can affect similarity estimates.

\section{Conclusion}
\label{sec:conclusion}

We introduced \ourmethod{}, a unified hierarchical feature-caching framework for training-free acceleration of image and video diffusion models. Starting from a redundancy analysis across intra-frame, inter-frame, motion, and denoising-step dimensions, \ourmethod{} coordinates Token Cache, Frame Cache, Block Cache, and Layered Cache within one budgeted inference-time system. Unlike averaging-based token merging baselines, \ourmethod{} uses similarity matching to select cacheable features and restores skipped positions from cached activations, preserving spatial-temporal order and positional consistency.

Experiments on SVD-XT, Latte, and SD3-medium show that \ourmethod{} consistently reduces inference latency by 25\%--35\% while maintaining visual fidelity and motion coherence. These results suggest that multidimensional cache-based feature reuse is a practical path toward more efficient high-resolution image and video generation without model retraining.

\appendix
\section*{Appendix}
\addcontentsline{toc}{section}{Appendix}

\section{Triton Kernel Acceleration}
\label{appendix:triton}
To reduce the overhead of cache selection and restoration, we replace the original PyTorch implementation with custom Triton kernels. These kernels optimize the gather--scatter pattern used to compact retained tokens, skip redundant positions, and restore cached activations to their original locations. The implementation uses block-level parallelism over batched token indices, coalesced memory access, and fused permutation/scatter operations to minimize global memory traffic. This is important because cache-based acceleration only improves end-to-end latency when the bookkeeping overhead is substantially smaller than the computation saved by skipping redundant features.

\subsection{Performance Evaluation of Triton Kernels}
\label{appendix:triton_results}

\begin{figure}[H]
\centering
\includegraphics[width=0.9\linewidth]{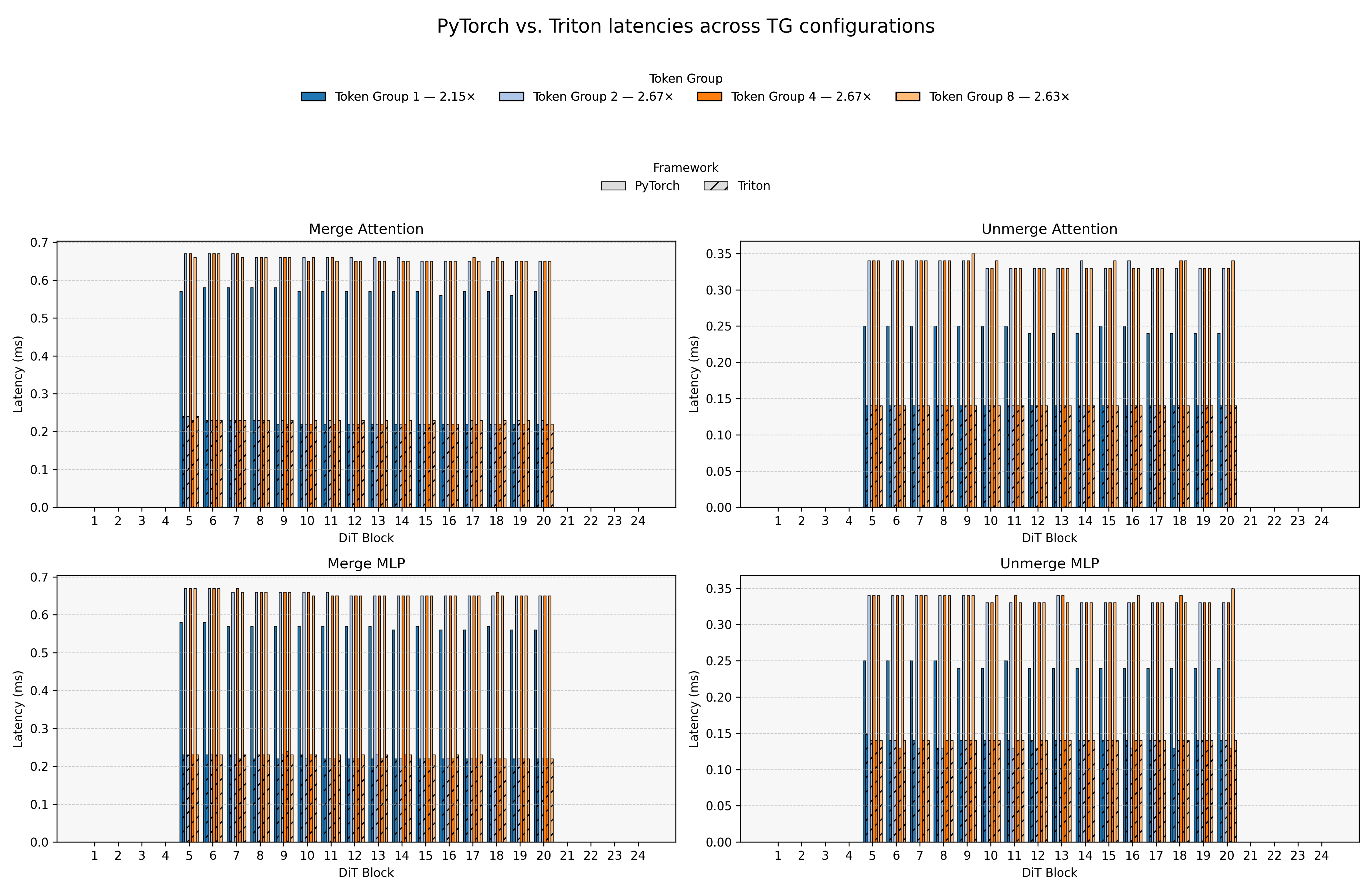}
\caption{Comparison of our custom Triton kernel implementation against the PyTorch baseline on SD3-medium under different token group counts in hierarchical caching. The Triton kernels achieve higher throughput and lower latency through optimized memory access and fused gather--scatter operations. Average speedups are shown alongside the token-group legend.}
\label{fig:triton_perf}
\end{figure}

We compare our custom Triton kernels against the baseline PyTorch implementation on SD3-medium under different token group counts ($1,2,4,8$) in the hierarchical caching pipeline. As described in Section~\ref{sec:design_choices}, we omit the initial and final four transformer blocks because they are sensitive to hierarchical caching and can affect generated image quality. As shown in Figure~\ref{fig:triton_perf}, the Triton kernels consistently outperform their PyTorch counterparts, achieving more than $2.0\times$ speedup for the selection and restoration routines and scaling better as the number of token groups increases.

\section{Cache Strategy Refinement Algorithm}
\label{appendix:refinement}

For completeness, we include the full pseudocode of the cache strategy refinement procedure introduced in Section~\ref{sec:method_cache_proposal}. This algorithm refines the preliminary cache schedule from Algorithm~\ref{alg:layer_cache} by pruning redundant cache steps, realigning dependencies, and optimizing cache intervals for computational efficiency.

\begin{algorithm}[H]
\caption{Cache Strategy Refinement Algorithm}\label{alg:ref_alg_detail}
\begin{algorithmic}
    \Require Model Compute Graph $\mathcal{G}$, CacheStrategy $\mathcal{C}$

    \State \textbf{Preprocess:}
    \For{each node in cache strategy $\mathcal{C}$}
        \State Initialize metadata (children, parents, flags, counters) for the node
    \EndFor

    \State \textbf{Build Dependency Graph:}
    \For{each node}
        \State Add child and parent relationships based on model structure
    \EndFor

    \State \textbf{Calculate Dependency Counters:}
    \For{each node}
        \State Set child\_num and parent\_num based on relationships
    \EndFor

    \State \textbf{Remove Redundant Cache Steps:}
    \For{each node in topological order}
        \State Use flags to identify and remove unnecessary cache steps
        \State Update flags for child nodes
    \EndFor

    \State \textbf{Remove Consecutive Cache Steps with Identical Outputs:}
    \For{each eligible node in reverse topological order}
        \For{each pair of consecutive cache steps}
            \If{outputs are identical}
                \State Remove the redundant cache step
            \EndIf
        \EndFor
    \EndFor

    \State \textbf{Recalculate Dependency Counters}

    \State \textbf{Align Cache Steps with Parent Dependencies:}
    \For{each node in topological order}
        \For{each cache step not in interval}
            \State Update to minimum parent cache step $\geq$ current step
        \EndFor
    \EndFor

    \State \textbf{Return} Optimized CacheStrategy $\hat{\mathcal{C}}$
\end{algorithmic}
\end{algorithm}

\noindent
This refinement process enforces structural consistency within the model's compute graph, eliminating redundant computations while preserving temporal and dependency integrity across cached layers.

\section{Broader Societal Impact}

\ourmethod{} improves accessibility, sustainability, and deployment efficiency for existing generative systems by reducing inference cost without retraining.

\paragraph{Energy Efficiency and Environmental Impact.}
Inference-time acceleration can reduce energy use for a fixed hardware configuration. For example, in our SVD-XT experiments, generating a 25-frame $576\times1024$ video takes approximately 110 seconds on the baseline and 82.4 seconds with \ourmethod{}. On a 300\,W GPU, this corresponds to roughly 9.2\,Wh and 6.9\,Wh per generation, respectively, or about 25\% lower energy use. At large deployment scales, this kind of per-sample reduction can translate into meaningful aggregate savings, though exact values depend on hardware utilization, datacenter efficiency, and energy sources.

\paragraph{Accessibility and Cost Reduction.}
Because \ourmethod{} is training-free and operates entirely at inference time, it can be applied to pretrained diffusion models without additional fine-tuning. This lowers computational barriers for researchers, small studios, and individual creators. For cloud services, reduced latency and GPU-hour consumption may also reduce operational cost.

\end{document}